\documentclass[11pt]{article}
\usepackage[hyperref]{acl2019}
\usepackage{adjustbox}
\usepackage{times}
\usepackage{url}
\usepackage{multirow}
\usepackage{graphicx}
\usepackage{booktabs}
\usepackage{enumitem}
\usepackage{array}
\aclfinalcopy
\newcommand{\club}{\ensuremath\clubsuit}
\newcommand{\spade}{\ensuremath\spadesuit}
\usepackage{makecell}
\title{A Comparison of Word-based and Context-based Representations for Classification Problems in Health Informatics}
\usepackage{booktabs}
\author{\begin{tabular}{ccc}
\multicolumn{3}{c}{Aditya Joshi$^{\spade}$, Sarvnaz Karimi$^{\spade}$, Ross Sparks$^{\spade}$, C\'ecile Paris$^{\spade}$, C Raina MacIntyre$^{\club}$}\\
\end{tabular}\\
\begin{tabular}{ccc}
\multicolumn{3}{c}{$^{\spade}$CSIRO Data61, Sydney, Australia}\\
\multicolumn{3}{c}{$^{\club}$Kirby Institute, University of New South Wales, Sydney, Australia}\\
\multicolumn{3}{c}{\{firstname.lastname\}@csiro.au , r.macintyre@unsw.edu.au}
\end{tabular}
}
\begin{document}
\maketitle
\begin{abstract}
Distributed representations of text can be used as features when training a statistical classifier. These representations may be created as a composition of word vectors or as context-based sentence vectors. We compare the two kinds of representations (word versus context) for three classification problems: influenza infection classification, drug usage classification and personal health mention classification. For statistical classifiers trained for each of these problems, context-based representations based on ELMo, Universal Sentence Encoder, Neural-Net Language Model and FLAIR are better than Word2Vec, GloVe and the two adapted using the MESH ontology. There is an improvement of 2-4\% in the accuracy when these context-based representations are used instead of word-based representations.
\end{abstract}

\section{Introduction}
Distributed representations (also known as `embeddings') are dense, real-valued vectors that capture semantics of concepts~\cite{word2vec}. When learned from a large corpus, embeddings of related words are expected to be closer than those of unrelated words. When a statistical classifier is trained, distributed representations of textual units (such as sentences or documents) in the training set can be used as feature representations of the textual unit. This technique of statistical classification that uses embeddings as features has been shown to be useful for many Natural Language Processing (NLP) problems~\cite{S15-2094,D16-1104,W16-4910,R17-1088,S16-1098,W17-5236} and biomedical NLP problems~\cite{E17-1109,U16-1003}. In this paper, we experiment with three classification problems in health informatics: influenza infection classification, drug usage classification and personal health mention classification. We use statistical classifiers trained on tweet vectors as features. To compute a tweet vector, \textit{i.e.}, a distributed representation for tweets, typical alternatives are: (a) tweet vector as a function of word embeddings of the content words\footnote{{\em Content words} refers to all words except stop words.} in the tweet; or, (b) a contextualised representation that computes sentence vectors using language models. The former considers meanings of words in isolation, while the latter takes into account the order of these words in addition to their meaning. We compare word-based and context-based representations for the three classification problems. This paper investigates the question:

\begin{quote}`\textit{When statistical classifiers are trained on vectors of tweets for health informatics, how should the vector be computed: using word-based representations that consider words in isolation or context-based representations that account for word order using language models?}'
\end{quote}

For these classification problems, we compare five approaches that use word-based representations with four approaches that use context-based representations. 
\begin{table*}[h!]
\centering
\begin{tabular}{|m{1em}|p{5.2cm}|p{8.3cm}|}
\hline
& \textbf{Representation}
 & \textbf{Details} \\ \hline
& \multicolumn{2}{c|}{\underline{A tweet vector is the average of the vectors of the content words in the tweet.}}\\
\multirow{2}{*}{\rotatebox{90}{\textbf{Word-based}}}& Word2Vec\_PreTrain, GloVe\_PreTrain & Vectors of the content words are obtained from pre-trained embeddings from Word2Vec \& GloVe respectively. \\
& Word2Vec\_SelfTrain & Vectors of the content words are based on embeddings learned from the training set, separately for each fold. \\
& Word2Vec\_WithMeSH, Glove\_WithMeSH & Vectors of the content words are pre-trained word embeddings from Word2Vec \& GloVe (respectively) retrofitted using MeSH ontology.\\
\hline
\multirow{2}{*}{\rotatebox{90}{\textbf{\small Context-based }}}& \multicolumn{2}{c|}{\underline{A tweet vector is obtained from a pre-trained language model that uses context.}}\\
& ELMo, USE, NNLM, FLAIR & Context-based representations of tweets are obtained from pre-trained models of ELMo, USE, NNLM and FLAIR respectively. They account for relationship between words using language models.\\
\hline
\end{tabular}
\caption{Summary of the representations used in our experiments.\label{tab:summary}}
\end{table*}

\section{Related Work}
Distributed representations as features for statistical classification have been used for many NLP problems: semantic relation extraction~\cite{hashimoto2015task}, sarcasm detection~\cite{D16-1104}, sentiment analysis~\cite{S15-2094,P18-1112}, co-reference resolution~\cite{R17-1088}, grammatical error correction~\cite{W16-4910}, emotion intensity determination~\cite{W17-5236} and sentence similarity detection~\cite{S16-1098}.  In terms of the biomedical domain, word embedding-based features have been used for entity extraction in biomedical corpora~\cite{E17-1109} or clinical information extraction~\cite{U16-1003}. Several approaches for personal health mention classification have been reported~\cite{D11-1145,lamb2013separating,yin2015scalable}. \citet{D11-1145} use bag-of-words as features for personal health mention classification. \citet{lamb2013separating} use linguistic features including coarse topic-based features, while \citet{yin2015scalable} use features based on parts-of-speech and dependencies for a statistical classifier. \citet{feng2018detecting} compare statistical classifiers with deep learning-based classifiers for personal health mention detection. In terms of detecting drug-related content in text, there has been work on detecting adverse drug reactions~\cite{karimi2015text}. \citet{nikfarjam2015pharmacovigilance} use word embedding clusters as features for adverse drug reaction detection.
\section{Representations}
A tweet vector is a distributed representation of a tweet, and is computed for every tweet in the training set. The tweet vector along with the output label is then used to train the statistical classification model. The intuition is that the tweet vector captures the semantics of the tweet and, as a result, can be effectively used for classification. To obtain tweet vectors, we experiment with two alternatives that have been used for several text classification problems in NLP: word-based representations and context-based representations. They are summarised in Table~\ref{tab:summary}, and described in the following subsections.

\subsection{Word-based Representations}
A word-based representation of a tweet combines word embeddings of the content words in the tweet. We use the average of the word embeddings of content words in the tweet. Average of word embeddings have been used for different NLP tasks~\cite{de2016representation,W18-4002, W18-4414,Y15-1013,C18-1152}. As in past work, words that were not learned in the embeddings are dropped during the computation of the tweet vector. We experiment with three kinds of word embeddings:
\begin{enumerate}
    \item \textbf{Pre-trained Embeddings}: Denoted as {\em Word2Vec\_PreTrained} and {\em GloVe\_PreTrained} in Table~\ref{tab:summary}, we use pre-trained embeddings of words learned from large text corpora: (A) Word2Vec by \citet{word2vec}: This has been pre-trained on a corpus of news articles with 300 million tokens, resulting in 300-dimensional vectors; (B) GloVe by \citet{glove}: This has been pre-trained on a corpus of tweets with 27 billion tokens, resulting in 200-dimensional vectors.
\item \textbf{Embeddings Trained on The Training Split}: It may be argued that, since the pre-trained embeddings are learned from a corpus from an unrelated domain (news and general, in the case of Word2Vec and GloVe respectively), they may not capture the semantics of the domain of the specific classification problem. Therefore, we also use the Word2Vec Model available in the gensim library~\cite{gensim} to learn word embeddings from the documents. For each split, the corresponding training set is used to learn the embeddings. The embeddings are then used to compute the tweet vectors and train the classifier. We refer to these as {\em Word2Vec\_SelfTrain}.
\item \textbf{Pre-trained embeddings retrofitted with medical ontologies}: Another alternative to adapt word embeddings for a classification problem is to use structured resources (such as ontologies) from a domain same as that of the classification problem. \citet{retrofitting} show that word embeddings can be retrofitted to capture relationships in an ontology. We use the Medical Subject Headings (MeSH) ontology~\cite{mesh}, maintained by the U.S. National Library of Medicine, which provides a hierarchically-organised terminology of medical concepts. Using the algorithm by ~\citet{retrofitting}, we retrofit pre-trained embeddings from Word2Vec and GloVe, with the MeSH ontology. The retrofitted embeddings for Word2Vec and GloVe are referred to as {\em Word2Vec\_WithMeSH}, and {\em GloVe\_WithMeSH} respectively.
\end{enumerate}
\begin{table}[]
\centering
\begin{tabular}{ll} 
\toprule
\textbf{Classification} & \textbf{\# tweets (\# true tweets)}\\ \midrule
IIC   & \hphantom{x}9,006 (2,306)    \\
DUC   & 13,409 (3,167) \\
PHMC  & \hphantom{x}2,661 (1,304) \\ \bottomrule
\end{tabular}
\caption{Dataset statistics.\label{tab:stats}}
\end{table}
The three kinds of word-based representations result in five configurations: {\em Word2Vec\_PreTrained}, {\em GloVe\_PreTrained},  {\em Word2Vec\_SelfTrain}, {\em Word2Vec\_WithMeSH}, and {\em GloVe\_WithMeSH}.
\subsection{Context-based Representations}
\begin{table*}[t]
\centering
\begin{tabular}{lcccc}
\toprule
 & \# dim. & \textbf{IIC} & \textbf{DUC} & \textbf{PHMC} \\ \midrule
\multicolumn{5}{c}{(A) \textit{Word-based Representations}} \\ \cmidrule{1-5} 
Word2Vec\_PreTrain & 300 & 0.8106 ($\sigma$: 0.024) & 0.7417 ($\sigma$: 0.153) & 0.7632 ($\sigma$: 0.037) \\ 
GloVe\_PreTrain & 200 & 0.7996 ($\sigma$: 0.015) & 0.7549 ($\sigma$: 0.120) & 0.7765 ($\sigma$: 0.033) \\ 
Word2Vec\_SelfTrain & 300 & 0.5099 ($\sigma$: 0.001) & 0.7450 ($\sigma$: 0.028) & 0.7418 ($\sigma$: 0.003)\\
Word2Vec\_WithMeSH & 300 & 0.6944 ($\sigma$: 0.021) & 0.7450 ($\sigma$: 0.046) & 0.7427 ($\sigma$: 0.050) \\
GloVe\_WithMeSH & 200 & 0.7264 ($\sigma$: 0.017) & 0.7635 ($\sigma$: 0.030) & 0.7425 ($\sigma$: 0.010) \\ 
\midrule
\multicolumn{5}{c}{(B) \textit{Context-based Representations}} \\ \cmidrule{1-5}  
 ELMo & 1024 & 0.8010 ($\sigma$: 0.021) & 0.7724 ($\sigma$: 0.090) & 0.7814 ($\sigma$: 0.02) \\ 
 USE & 512 & 0.8164 ($\sigma$: 0.008) & \textbf{0.7790} ($\sigma$: 0.100) & \textbf{0.8155} ($\sigma$: 0.030) \\
 NNLM & 128 & \textbf{0.8520} ($\sigma$: 0.006) & 0.7610 ($\sigma$: 0.070) & 0.7495 ($\sigma$: 0.020)\\
% BERT & & 0.7129 ($\sigma$:0.014) & 0.7416 ($\sigma$: 0059) & 0.7416 ($\sigma$: 0.011) \\ % bert_uncased_L-24_H-1024_A-16
FLAIR & 4196 & 0.8000 ($\sigma$: 0.021) & 0.7667 ($\sigma$: 0.116) & 0.7896 ($\sigma$: 0.031)\\
\bottomrule
\end{tabular}
\caption{Comparison of five word-based representations with four context-based representations; Average accuracy with standard deviation ($\sigma$) indicated in brackets.\label{tab:res}}
\end{table*}
%\begin{table*}[t]
%\centering
%\begin{tabular}{|p{4cm}|p{3.5cm}|p{3.5cm}|p{3.5cm}|}
%\hline
% & \textbf{IIC} & \textbf{DUC} & \textbf{PHMC} \\ \hline
% \multicolumn{4}{|c|}{\textit{Word-based Representations}} \\ \hline 
% \multicolumn{4}{|c|}{\textbf{Pre-trained embeddings}} \\ \hline
%Word2Vec\_PreTrain & 0.8106 ($\sigma$: 0.024) & 0.7417 ($\sigma$: 0.153) & 0.7632 ($\sigma$: 0.037) %\\ 
%GloVe\_PreTrain & 0.7996 ($\sigma$: 0.015) & 0.7549 ($\sigma$: 0.120) & 0.7765 ($\sigma$: 0.033) \\ \hline
%\multicolumn{4}{|c|}{\textbf{Embeddings Trained on Training Corpus}}\\ \hline
%Word2Vec\_SelfTrain & 0.5099 ($\sigma$: 0.001) & 0.7450 ($\sigma$: 0.028) & 0.7418 ($\sigma$: 0.003)\\ \hline
%\multicolumn{4}{|c|}{\textbf{Pre-trained embeddings retrofitted with MeSH}}\\ \hline
%Word2Vec\_WithMeSH & 0.6944 ($\sigma$: 0.021) & 0.7450 ($\sigma$: 0.046) & 0.7427 ($\sigma$: 0.050) \\
%GloVe\_WithMeSH & 0.7264 ($\sigma$: 0.017) & 0.7635 ($\sigma$: 0.030) & 0.7425 ($\sigma$: 0.010) \\ \hline
% \multicolumn{4}{|c|}{\textit{Context-based Representations}} \\ \hline
% ELMo & 0.8010 ($\sigma$: 0.021) & 0.7724 ($\sigma$: 0.090) & 0.7814 ($\sigma$: 0.02) \\ 
% USE & 0.8164 ($\sigma$: 0.008) & \textbf{0.7790} ($\sigma$: 0.100) & \textbf{0.8155} ($\sigma$: 0.030) \\
% NNLM & \textbf{0.8520} ($\sigma$: 0.006) & 0.7610 ($\sigma$: 0.070) & 0.7495 ($\sigma$: 0.020)\\
%\hline
%\end{tabular}
%\caption{Comparison of five word-based representations with three context-based representations; Average 10-fold accuracy with standard deviation ($\sigma$) indicated in brackets.\label{tab:res}}
%\end{table*}

Context-based representations may use language models to generate vectors of sentences. Therefore, instead of learning vectors for individual words in the sentence, they compute a vector for sentences on the whole, by taking into account the order of words and the set of co-occurring words.

We experiment with four deep contextualised vectors: (A) \textbf{Embeddings from Language Models (ELMo)} by \citet{elmo}: ELMo uses character-based word representations and bidirectional LSTMs. The pre-trained model computes a contextualised vector of 1024 dimensions.
    ELMo is available in the Tensorflow Hub\footnote{\url{https://www.tensorflow.org/hub/}; Accessed on 3rd June, 2019.}, a repository of machine learning modules; (B) \textbf{Universal Sentence Encoder (USE)} by \citet{use}: The encoder uses a Transformer architecture that uses attention mechanism to incorporate information about the order and the collection of words~\cite{vaswani2017attention}. The pre-trained model of USE that returns a vector of 512 dimensions is also available on Tensorflow Hub; (C) \textbf{Neural-Net Language Model (NNLM)} by \citet{nnlm}: The model simultaneously learns representations of words and probability functions for word sequences, allowing it to capture semantics of a sentence. We use a pre-trained model available on Tensorflow Hub, that is trained on the English Google News 200B corpus, and computes a vector of 128 dimensions; (D) \textbf{FLAIR} by \citet{flair}: This library by Zalando research\footnote{\url{https://github.com/zalandoresearch/flair}; Accessed on 3rd June, 2019.} uses character-level language models to learn contextualised representations. We use the pooling option to create sentence vectors. This is a concatenation of GloVe embeddings and the forward/backward language model. The resultant is a vector of 4196 dimensions.

Table~\ref{tab:summary} refers to the four configurations as {\em ELMo}, {\em USE}, {\em NNLM} and {\em FLAIR} respectively.
\section{Experiment Setup}
We conduct our experiments on three boolean classification problems in health informatics: (A) \textbf{Influenza Infection Classification (IIC)}: The goal is to predict if a tweet reports an influenza infection (`\textit{I have been coughing all day}', for example) or describes information about influenza (`\textit{flu outbreaks are common in this month of the year}', for example). We use the dataset presented in~\citet{lamb-paul-dredze-naacl-2013}; (B) \textbf{Drug Usage Classification (DUC)}: The objective here is to detect whether or not a tweet describes the usage of a medicinal drug (`\textit{I took some painkillers this morning}', for example). We use the dataset provided by ~\citet{jiang2016construction}; (C) \textbf{Personal Health Mention classification (PHMC)}: A personal health mention is a person's report about their illness. We use the dataset provided by~\citet{bella2015}. For example `\textit{I have been sick for a week now}' is a personal health mention while `\textit{Rollercoasters can make you sick}' is not. It must be noted that IIC involves influenza while the PHMC dataset covers a set of illnesses as described later. 

The datasets for each of the classification problems consist of tweets that have been manually annotated as reported in the corresponding papers. The statistics of these datasets are shown in Table~\ref{tab:stats}. The values in brackets indicate the number of true tweets (\textit{i.e.}, tweets that have been labeled as true), since these are boolean classification problems. For details on inter-annotator agreement and the annotation techniques, we refer the reader to the original papers. Based on sentence vectors obtained using either word-based or context-based representations, we train logistic regression with default parameters available as a part of the Liblinear package~
\cite{fan2008liblinear}. We report five-fold cross-validation results for our experiments. Each fold is created using stratified k-fold sampling available in scikit-learn\footnote{\url{https://scikit-learn.org/stable/}; Accessed on 3rd June, 2019.}.

\section{Results}
We first present a quantitative evaluation to compare the two types of representations. Following that, we analyse sources of errors.
\subsection{Quantitative Evaluation}
We compare word-based and context-based representations for the three classification problems in Table~\ref{tab:res}. Accuracy is computed as the proportion of correctly classified instances. The table contains the average accuracy values with standard deviation values shown in parentheses. The table is divided into two parts. Part (A) corresponds to experiments using word-based representations, while Part (B) corresponds to those using context-based representations. In general, context-based representations result in an improvement in the three classification problems as compared to word-based representations. For IIC, the best word-based representation is when pre-trained Word2Vec embeddings ($Word2Vec\_PreTrain$) of content words are averaged to generate the tweet vector. The accuracy in this case is 0.8106. In contrast, the best performing context-based representation is NNLM (0.8520). This is an improvement of 4\% points. Similarly, tweet vectors created using USE result in an accuracy of 0.7790 for DUC and 0.8155 for PHMC. This is an improvement of 2-4\% points each over the word-based representations for these two classification problems as well. In addition, for pre-trained embeddings (Word2Vec and GloVe) retrofitted with a medical ontology (MeSH), we observe a degradation in the accuracy for IIC and PHMC, as compared to without retrofitting. There is an improvement of 1\% point in the case of DUC. Similarly, learning the embeddings on the specific training corpus does not work well. It leads to a degradation as compared to pre-trained embeddings. This could happen because pre-trained embeddings are trained on much larger corpora than our training datasets, thereby capturing semantics more effectively than the \textit{Word2Vec\_SelfTrain} variant.

\begin{table}[]
    \centering
    \begin{tabular}{p{1cm}cccc}
    \toprule
     & \multicolumn{2}{p{0.36\linewidth}}{\textbf{$1^{st}$-person mentions}}& \multicolumn{2}{p{0.36\linewidth}}{\textbf{Present Participle}}\\ \midrule
 & \textbf{Word} & \textbf{Context} & \textbf{Word} & \textbf{Context} \\ \midrule
        IIC & 58.2 & \textbf{41.0}  & 79.6 & \textbf{72.5}\\
        DUC & 66.4 & \textbf{54.75} & \textbf{33.0} & 40.75 \\
        PHMC& 64.8 & \textbf{37.5} & 61.6 & \textbf{40.0}\\
        \bottomrule
    \end{tabular}
    \caption{Average number of instances (out of 100 randomly sampled mis-classified instances) containing first-person mentions and present participle form for the three classification problems and two types of representations.\label{tab:erroranal}}
\end{table}

\subsection{Qualitative Evaluation}
For a qualitative comparison of the two representations, we analyse 100 randomly sampled instances that are mis-classified by each classifier. While these instances need not be the same for each classifier, the trends in the errors show where one kind of representation scores over the other. We compared linguistic properties of these mis-classified instances, such as the person, tense and number. Table~\ref{tab:erroranal} shows two linguistic properties where we observed the most variation: first-person mentions and the use of present participles. The two properties are important in terms of the semantics of the three classification problems. First-person mentions are useful indicators to identify if the speaker has influenza, took a drug or reported a personal health mention. Similarly, present participle forms of verbs appear in situations where a person has had an infection or taken a drug. For `Word', the average is over the five representations, while for `Context', the average is over the four context-based representations. In the case of IIC, an average of 58.2 mis-classified instances from word-based representations contained first person mentions. The corresponding number for context-based representations was 41. For PHMC, the averages are 64.8 (word-based) and 37.5 (context-based). The difference is not as high in the case of DUC  (66.4 and 54.75 respectively). Differences are observed in the case of present participle in mis-classified instances. However, in the case of DUC, errors from context-based representations contain more average number of present participles (40.75) than word-based representations (33).

\section{Conclusions}
In this paper, we show that context-based representations are a better choice than word-based representations to create tweet vectors for classification problems in health informatics. We experiment with three such problems: influenza infection classification, drug usage classification and personal health mention classification, and compare word-based representations with context-based representations as features for a statistical classifier. For word-based representations, we consider pre-trained embeddings of Word2Vec and GloVe, embeddings trained on the training split, and the pre-trained embeddings of Word2Vec and GloVe retrofitted to a medical ontology. For context-based representations, we consider ELMo, USE, NNLM and FLAIR. For the three problems, the highest accuracy is obtained using context-based representations. In comparison with pre-trained embeddings, the improvement in classification is approximately 4\% for influenza infection classification, 2\% for drug usage classification and 4\% for personal health mention classification. Embeddings trained on the training corpus or retrofitted on the ontology perform worse than those pre-trained on a large corpus. 

While these observations are based on statistical classifiers, the corresponding benefit of context-based representations on neural architectures can be validated as a future work. In addition, while we average the word vectors to obtain tweet vectors, other options for tweet vector computation can be considered for word-based representations. In terms of the dataset, the comparison should be validated for text forms other than tweets, such as medical records. Medical records are expected to have typical challenges such as the use of abbreviations and domain-specific phrases that may not have been learned in pre-trained embeddings.
\section*{Acknowledgment}
The authors would like to thank the anonymous reviewers for their helpful comments.
\bibliography{biblioo}
\bibliographystyle{acl_natbib}
\end{document}